%% file: formatting-instructions-latex-2019.tex
\def\year{2019}\relax
\documentclass[letterpaper]{article} 
\usepackage{aaai19}  
\usepackage{times}  
\usepackage{helvet}  
\usepackage{courier}  
\usepackage{url}  
\usepackage{graphicx}  
\usepackage{amsmath}
\usepackage{amssymb}
\usepackage{subcaption}
\usepackage[export]{adjustbox}
\usepackage{enumitem}
\setlist[itemize]{leftmargin=*}
\usepackage{natbib}
\usepackage[breaklinks=true,bookmarks=false]{hyperref}

\input{abbrev}

\begin{document}

%
\title{Large-Scale Visual Relationship Understanding}
\author{
Ji Zhang\textsuperscript{1,2},
Yannis Kalantidis\textsuperscript{1},
Marcus Rohrbach\textsuperscript{1},\\
\bf \Large Manohar Paluri\textsuperscript{1},
Ahmed Elgammal\textsuperscript{2}, 
 Mohamed Elhoseiny\textsuperscript{1} \\
\\
\textsuperscript{1}Facebook Research  \\
\textsuperscript{2}Department of Computer Science, Rutgers University}
\maketitle

\input{abstract}

\input{introduction}

\input{related}

\input{method}

\input{experiments}

\input{conclusions}


{\small
\bibliographystyle{aaai}
\bibliography{egbib}
}

\end{document}

%% file: abbrev.tex
\newcommand{\vx}{\textbf{x}}
\newcommand{\vy}{\textbf{y}}

\newcommand{\sregion}{\mathcal{R_S}}
\newcommand{\pregion}{\mathcal{R_P}}
\newcommand{\oregion}{\mathcal{R_O}}

 \usepackage{xspace}
 \makeatletter
 \DeclareRobustCommand\onedot{\futurelet\@let@token\@onedot}
 \def\@onedot{\ifx\@let@token.\else.\null\fi\xspace}
 \def\eg{e.g\onedot} 
 \def\ie{\emph{i.e}\onedot}

 \makeatother

\usepackage[dvipsnames]{xcolor}

\newcommand{\rel}[1]{$\langle$#1$\rangle$}

\newcommand{\head}[1]{{\noindent \textbf{#1}}}

%% file: abstract.tex
\begin{abstract}
Large scale visual understanding is challenging, as it requires a model to handle the widely-spread and imbalanced distribution of \rel{subject, relation, object} triples. In real-world scenarios with large numbers of objects and relations, some are seen very commonly while others are barely seen. We develop a new relationship detection model that embeds objects and relations into two vector spaces where both discriminative capability and semantic affinity are preserved. We learn a visual and a semantic module that map features from the two modalities into a shared space, where matched pairs of features have to discriminate against those unmatched, but also maintain close distances to semantically similar ones. Benefiting from that, our model can achieve superior performance even when the visual entity categories scale up to more than $80,000$, with extremely skewed class distribution. We demonstrate the efficacy of our model on a large and imbalanced benchmark based of Visual Genome that comprises $53,000+$ objects and $29,000+$ relations, a scale at which no previous work has been evaluated at. We show superiority of our model over competitive baselines on the original Visual Genome dataset with $80,000+$ categories. We also show state-of-the-art performance on the VRD dataset and the scene graph dataset which is a subset of Visual Genome with $200$ categories.
\end{abstract}


%% file: introduction.tex
\section{Introduction}
\label{sec:introduction}

\begin{figure}[t!]
\centering
\includegraphics[trim={0 0 0 19cm},clip,width=.98\linewidth]{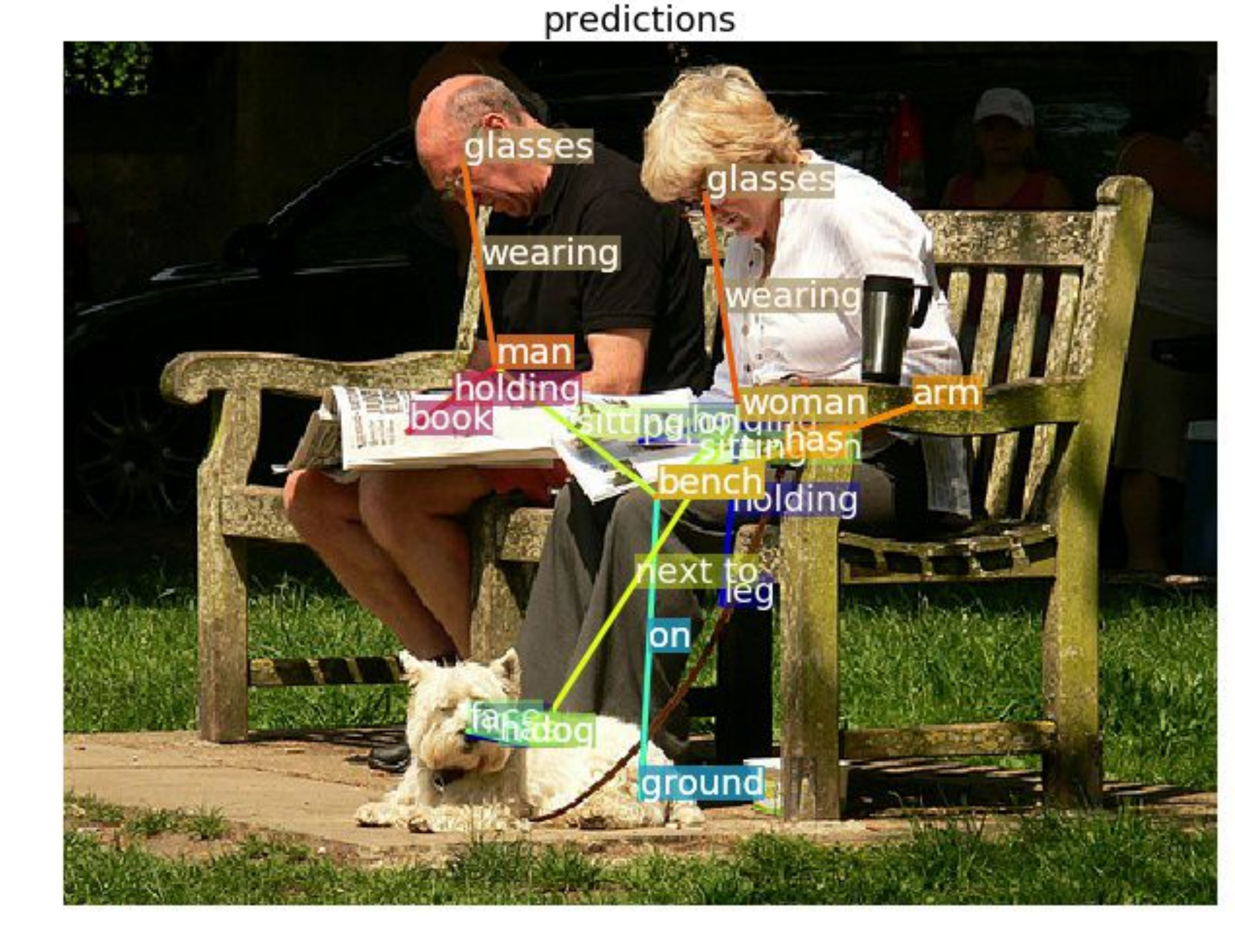}
\setlength\belowcaptionskip{-2ex}
\captionsetup{font=small}
\caption{Relationships predicted by our approach on an image. Different relationships are colored differently with a relation line connecting each subject and object. Our model is able to recognize relationships composed of over $53,000$ object categories and over $29,000$ relation categories.}
\label{fig:teaser}
\end{figure}

Scale matters. In the real world, people tend to describe visual entities with open vocabulary, \eg, the raw ImageNet \citep{deng2009imagenet} dataset has 21,841 synsets that cover a vast range of objects. The number of entities is significantly larger for relationships since the combinations of $\langle$subject, relation, object$\rangle$ are orders of magnitude more than objects \citep{lu2016visual,plummerPLCLC2017,zhang2017relationship}. Moreover, the long-tailed distribution of objects can be an obstacle for a model to learn all classes sufficiently well, and such challenge is exacerbated in relationship detection because either the subject, the object, or the relation could be infrequent, or their triple might be jointly infrequent. Figure~\ref{fig:teaser} shows an example from the Visual Genome dataset, which contains commonly seen relationship (e.g., \rel{man,wearing,glasses}) along with uncommon ones (e.g., \rel{dog,next to,woman}).

Another challenge is that object categories are often semantically associated \citep{deng2009imagenet,krishna2017visual,deng2014large}, and such connections could be more subtle for relationships since they are conditioned on the contexts. For example, an image of \rel{person,ride,horse} could look like one of \rel{person,ride,elephant} since they both belong to the kind of relationships where a person is riding an animal, but \rel{person,ride,horse} would look very different from \rel{person,walk with,horse} even though they have the same subject and object. It is critical for a model to be able to leverage such conditional connections.

In this work, we study relationship recognition at an unprecedented scale where the total number of visual entities is more than 80,000. To achieve that we use a continuous output space for objects and relations instead of discrete labels.
 We demonstrate superiority of our model over competitive baselines on a large and imbalanced benchmark based of Visual Genome that comprises $53,000+$ objects and $29,000+$ relations. We also achieve state-of-the-art performance on the Visual Relationship Detection (VRD) dataset \citep{lu2016visual}, and the scene graph dataset \citep{xu2017scenegraph}.

%% file: related.tex
\section{Related Work}
\label{sec:related}
 



\noindent \textbf{Visual Relationship Detection}
A large number of visual relationship detection approaches have emerged during the last couple of years. Almost all of them are based on a small vocabulary, e.g., 100 object and 70 relation categories from the VRD dataset \citep{lu2016visual}, or a subset of VG with the most frequent object and relation categories \citep{zhang2017visual,xu2017scenegraph,zhang2018introduction,zhang2018interpretable,zhang2019CVPR}.

%

In one of the earliest works, \citet{lu2016visual} utilize the object detection output of an an R-CNN detector and leverage language priors from semantic word embeddings to fine-tune the likelihood of a predicted relationship.
Very recently, \citet{Zhuang_2017_ICCV} use language representations of the subject and object as ``context'' to derive a better classification result for the relation.
However, similar to \citet{lu2016visual} their language representations are pre-trained.
Unlike these approach, we fine-tune subject and object representations \textit{jointly} and employ the interaction between branches also at an earlier stage before classification. 

In \citet{yu17iccv}, the authors employ knowledge distillation from a large Wikipedia-based corpus and get state-of-the-art results for the VRD \citep{lu2016visual} dataset. In ViP-CNN \citep{LiCVPR2017}, the authors pose the problem as a classification task on limited classes and therefore cannot scale to the open-vocabulary scenarios. In our model we exploit co-occurrences at the relationship level to model such knowledge. Our approach directly targets the large category scale and is able to utilize semantic associations to compensate for infrequent classes, while at the same time achieves competitive performance in the smaller and constrained VRD \citep{lu2016visual} dataset.

Very recent approaches like \citet{zhao2017open,plummerPLCLC2017} target open-vocabulary for scene parsing and visual relationship detection, respectively. In \citet{plummerPLCLC2017}, the related work closest to ours, the authors learn a CCA model on top of different combinations of the subject, object and union regions and train a Rank SVM. They however consider each relationship triplet as a class and learn it as a whole entity, thus cannot scale to our setting.
Our approach embeds the three components of a relationship separately to the independent semantic spaces for object and relation, but implicitly learns connections between them via visual feature fusion and semantic meaning preservation in the embedding space.




\noindent \textbf{Semantically Guided Visual Recognition}.
Another parallel category of vision and language tasks is known as zero-shot/few-shot, where class imbalance is a primary assumption.
In \citet{NIPS13DeViSE}, \citet{norouzi2014zero} and \citet{NIPS13CMT}, word embedding language models (\eg, \citet{mikolov2013distributed}) were adopted to represent class names as vectors and hence allow zero-shot recognition. For fine-grained objects like birds and flowers, several works adopted Wikipedia Articles to guide zero-shot/few-shot recognition
\citep{ba2015predicting,elhoseiny2017sherlock}.
However, for relations and actions, these methods are not designed with the capability of locating the objects or interacting objects for visual relations. 
Several approaches have been proposed to model the visual-semantic embedding in the context of the image-sentence similarity task (\eg, \citet{kiros2014unifying,faghri2017vse++,Wang_2016_CVPR,gong2014multi}). Most of them focused on leaning semantic connections between the two modalities, which we not only aim to achieve, but with a manner that does not sacrifice discriminative capability since our task is detection instead of similarity-based retrieval. In contrast, visual relationship also has a structure of $\langle$subject, relation, object$\rangle$ and we show in our results that proper design of a visual-semantic embedding architecture and loss is critical for good performance.

Note: in this paper we use ``relation'' to refer to what is also known as `predicate'' in previous works, and ``relationship'' or ``relationship triplet'' to refer to a $\langle$subject, relation, object$\rangle$ tuple.

%% file: method.tex
\section{Method}
\label{sec:method}

\begin{figure*}[t!]
\centering
\adjincludegraphics[width=\linewidth,trim={0 {.07\height} 0 0},clip]{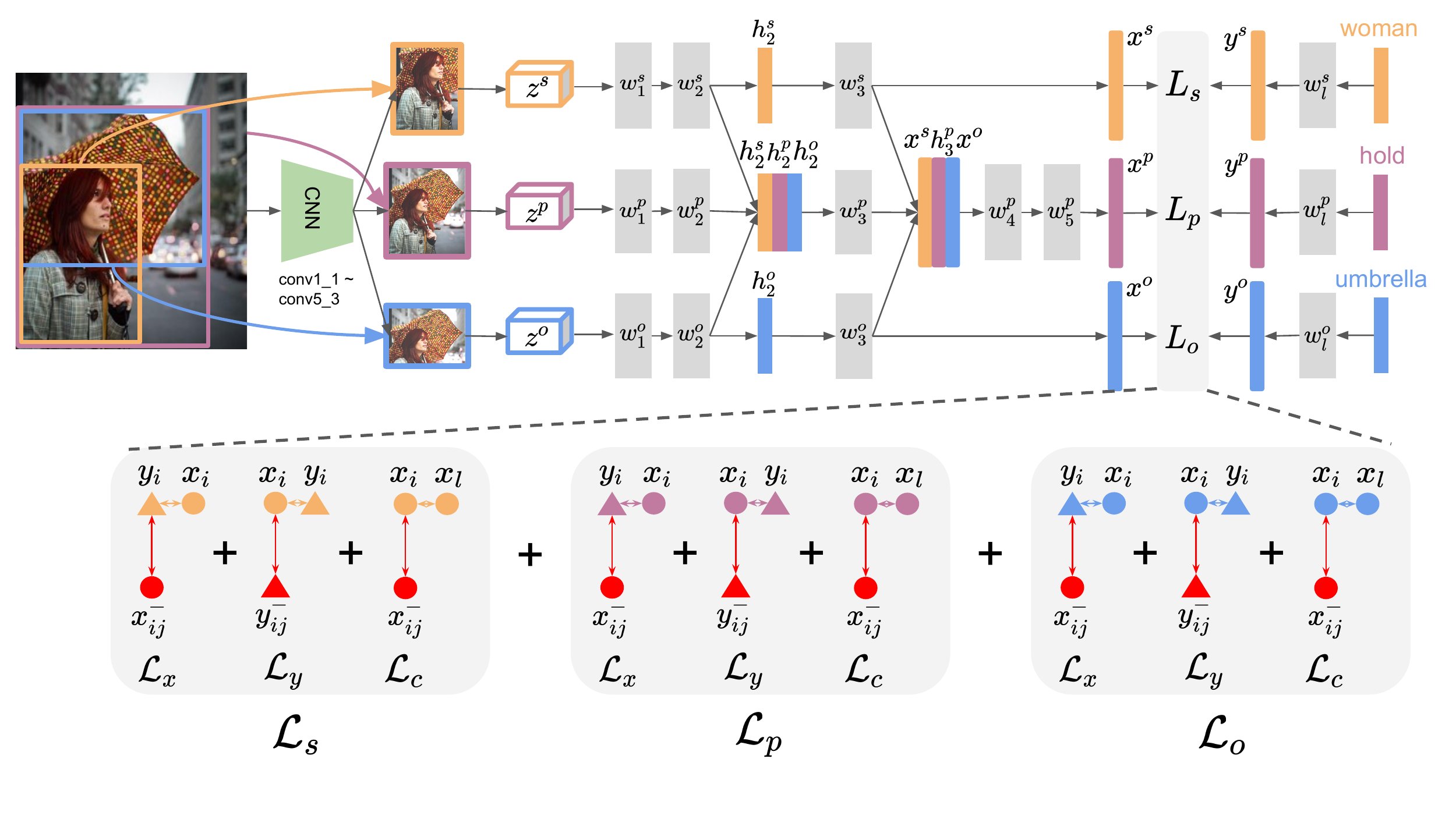}
\setlength\belowcaptionskip{-2ex}
\captionsetup{font=small}
\caption{(a) Overview of the proposed approach. $L_s$, $L_p$, $L_o$ are the losses of subject, relation and object. Orange, purple and blue colors represent subject, relation, object, respectively. Grey rectangles are fully connected layers, which are followed by ReLU activations except the last ones, \emph{\ie} $w^s_3$, $w^p_5$, $w^o_3$. We share layer weights of the subject and object branches, \emph{\ie} $w^s_i$ and $w^o_i$, $i=1,2...5$.}
\label{fig:model}
\end{figure*}

Figure \ref{fig:model} shows the work flow of our model. We take an image as input to the visual module and output three visual embeddings $x^s, x^p,$ and $x^o$ for subject, relation, and object. During training we take word vectors of subject, relation, object as input to the semantic module and output three semantic embeddings $y^s, y^p, y^o$. We minimize the loss by matching the visual and semantic embeddings using our designed losses. During testing we feed word vectors of all objects and relations and use nearest neighbor searching to predict relationship labels. The following sections describe our model in details.





\subsection{Visual Module}
\label{subsec:visual_module}
The design logic of our visual module is that a relation exists when its subject and object exist, but not vice versa. Namely, relation recognition is conditioned on subject and object, but object recognition is independent from relations. The main reason is that we want to learn embeddings for subject and object in a separate semantic space from the relation space. That is, we want to learn a mapping from visual feature space (which is shared among subject/object and relation) to the two separate semantic embedding spaces (for objects and relations). Therefore, involving relation features for subject/object embeddings would have the risk of entangling the two spaces.
Following this logic, as shown in Figure \ref{fig:model} an image is fed into a CNN ($conv1\_1$ to $conv5\_3$ of VGG16) to get a global feature map of the image, then the subject, relation and object features $z^s$, $z^p$, $z^o$ are ROI-pooled with the corresponding regions $\sregion$, $\pregion$, $\oregion$, each branch followed by two fully connected layers which output three intermediate hidden features $h^s_2$, $h^p_2$, $h^o_2$. For the subject/object branch, we add another fully connected layer $w^s_3$ to get the visual embedding $x^s$, and similarly for the object branch to get $x^o$. For the relation branch, we apply a two-level feature fusion: we first concatenate the three hidden features $h^s_2$, $h^p_2$, $h^o_2$ and feed it to a fully connected layer $w^p_3$ to get a higher-level hidden feature $h^p_3$, then we concatenate the subject and object embeddings $x^s$ and $x^o$ with $h^p_3$ and feed it to two fully connected layers $w^p_4$ $w^p_5$ to get the relation embedding $x^p$.



\subsection{Semantic Module}
\label{subsec:semantic_module}

On the semantic side, we feed word vectors of subject, relation and object labels into a small MLP 
of one or two $fc$ layers which outputs the embeddings. As in the visual module, the subject and object branches share weights while the relation branch is independent.
The purpose of this module is to map word vectors into an embedding space that is more discriminative than the raw word vector space while preserving semantic similarity. During training, we feed the ground-truth labels of each relationship triplet as well as labels of negative classes into the semantic module, as the following subsection describes; during testing, we feed the whole sets of object and relation labels into it for nearest neighbors searching among all the labels to get the top $k$ as our prediction.

A good word vector representation for object/relation labels is critical as it provides proper initialization that is easy to fine-tune on. We consider the following word vectors:

\head{Pre-trained word2vec embeddings (wiki).} We rely on the pre-trained word embeddings provided by \citet{mikolov2013distributed} which are widely used in prior work. We use this embedding as a baseline, and show later that by combining with other embeddings we achieve better discriminative ability.



\head{Relationship-level co-occurrence embeddings (relco).} We train a skip-gram word2vec model that tries to maximize classification of a word based on another word in the same context. As is in our case we define context via our training set's relationships, we effectively learn to maximize the likelihoods of $P(P|S,O)$ as well as $P(S|P,O)$ and $P(O|S,P)$.
Although maximizing $P(P|S,O)$ is directly optimized in \citet{yu17iccv}, we achieve similar results by reducing it to a skip-gram model and enjoy the scalability of a word2vec approach. 

\head{Node2vec embeddings (node2vec).} As the Visual Genome dataset further provides image-level relation graphs, we also experimented with training \textit{node2vec} embeddings as in \citet{grover2016node2vec}. These are effectively also word2vec embeddings, but the context is determined by random walks on a graph. In this setting, nodes correspond to subjects, objects and relations from the training set and edges are directed from $S \rightarrow P$ and from $P \rightarrow O$ for every image-level graph. This embedding can be seen as an intermediate between image-level and relationship level co-occurrences, with proximity to the one or the other controlled via the length of the random walks.

\subsection{Training Loss}
\label{subsec:our_loss}

To learn the joint visual and semantic embedding we employ a modified triplet loss. Traditional triplet loss \citep{kiros2014unifying} encourages matched embeddings from the two modalities to be closer than the mismatched ones by a fixed margin, while our version tries to maximize this margin in a softmax form. In this subsection we review the traditional triplet loss and then introduce our triplet-softmax loss in a comparable fashion. To this end, we denote the two sets of triplets for each positive visual-semantic pair by $(\vx^{l}, \vy^{l})$:
\begin{align}
tri^l_\vx &= \{ \vx^{l}, \vy^{l}, \vx^{l-}\} \\
tri^l_\vy &= \{ \vx^{l}, \vy^{l}, \vy^{l-}\} 
\label{eq:triplets}
\end{align}
where $l \in \{ s,p,o\}$, and the two sets $tri_\vx, tri_\vy$ correspond to triplets with negatives from the visual and semantic space, respectively. 

\head{Triplet loss.} If we omit the superscripts $\{s,p,o\}$ for clarity, the triplet loss $\mathcal{L}^{Tr}$ for each branch is summation of two losses $\mathcal{L}_{\vx}^{Tr}$ and $\mathcal{L}_{\vy}^{Tr}$:
\begin{align}
\mathcal{L}_{\vx}^{Tr} &= \frac{1}{NK} \sum_{i=1}^{N} \sum_{j=1}^{K} \max [0, m + s(\vy_i, \vx_{ij}^-) - s(\vy_i, \vx_i)] \\
\mathcal{L}_{\vy}^{Tr} &= \frac{1}{NK} \sum_{i=1}^{N} \sum_{j=1}^{K} \max [0, m + s(\vx_i, \vy_{ij}^-) - s(\vx_i, \vy_i)] \\
\mathcal{L}^{Tr} &= \mathcal{L}_{\vx}^{Tr} + \mathcal{L}_{\vy}^{Tr}
\label{eq:triplet_loss}
\end{align}
where $N$ is the number of positive ROIs, $K$ is the number of negative samples \textit{per positive} ROI, $m$ is the margin between the distances of positive and negative pairs, and $s(\cdot,\cdot)$ is a similarity function.

We can observe from Equation (3) that as long as the similarity between positive pairs is larger than that between negative ones by margin $m$, $[m + s(\vx_i, \vx_{ij}^-) - s(\vx_i, \vy_i)]\leq0$, and thus $max(0,\cdot)$ will return zero for that part. That means, during training once the margin is pushed to be larger than $m$, the model will stop learning anything from that triplet. Therefore, it is highly likely to end up with an embedding space where points are not discriminative enough for a classification-oriented task.

It is worth noting that although theoretically traditional triplet loss can pushes the margin as much as possible when $m=1$, most previous works (\eg, \citet{kiros2014unifying,faghri2017vse++,Gordo_2017_CVPR}) adopted a small $m$ to allow slackness during training. It is also unclear how to determine the exact value of $m$ given a specific task. We follow previous works and set $m=0.2$ in all of our experiments.


\head{Triplet-Softmax loss.} The issue of triplet loss mentioned above can be alleviated by applying softmax on top of each triplet, \emph{\ie}:
\begin{align}
\mathcal{L}_{\vx}^{TrSm} &= \frac{1}{N} \sum_{i=1}^{N} -\log \frac{e^{s(\vy_i, \vx_i)}}{e^{s(\vy_i, \vx_i)} + \sum_{j=1}^{K} e^{s(\vy_i, \vx_{ij}^-)}} \\
\mathcal{L}_{\vy}^{TrSm} &= \frac{1}{N} \sum_{i=1}^{N} -\log \frac{e^{s(\vx_i, \vy_i)}}{e^{s(\vx_i, \vy_i)} + \sum_{j=1}^{K} e^{s(\vx_i, \vy_{ij}^-)}} \\
\mathcal{L}^{TrSm} &= \mathcal{L}_{\vx}^{TrSm} + \mathcal{L}_{\vy}^{TrSm}
\label{eq:riplet_softmax_loss}
\end{align}
where $s(\cdot,\cdot)$
is the same similarity function (we use cosine similarity in this paper). All the other notations are the same as above. For each positive pair $(\vx_i, \vy_i)$ and its corresponding set of negative pairs $(\vx_i, \vy_{ij}^-)$, we calculate similarities between each of them and put them into a softmax layer followed by multi-class logistic loss so that the similarity of positive pairs would be pushed to be $1$, and $0$ otherwise. Compared to triplet loss, this loss always tries to enlarge the margin to its largest possible value (\emph{\ie}, 1), thus has more discriminative power than the traditional triplet loss.

\head{Visual Consistency loss.} To further force the embeddings to be more discriminative, we add a loss that pulls closer the samples from the same category while pushes away those from different categories, i.e.:
\begin{align}
\begin{split}
\mathcal{L}_{c} = \frac{1}{NK} \sum_{i=1}^{N} \sum_{j=1}^{K} \max [0, m + s(\vx_i, \vx_{ij}^-) - \min_{l\in\mathcal{C}(i)}s(\vx_i, \vx_l)]
\label{eq:losstriplet}
\end{split}
\end{align}
where $N$ is the number of positive ROIs, $\mathcal{C}(l)$ is the set of positive ROIs in the same class of $\vx_i$, $K$ is the number of negative samples \textit{per positive} ROI and $m$ is the margin between the distances of positive and negative pairs. The interpretation of this loss is: the minimum similarity between samples from the same class should be larger than any similarity between samples from different classes by a margin. Here we utilize the traditional triplet loss format since we want to introduce slackness between visual embeddings to prevent embeddings from collapsing to the class centers.

Empirically we found it the best to use triplet-softmax loss for $\mathcal{L}_{\vy}$ while using triplet loss for $\mathcal{L}_{\vx}$.
The reason is similar with that of the visual consistency loss: mode collapse should be prevented by introducing slackness.
On the other hand, there is no such issue for $y$ since each label $y$ is a mode by itself, and we encourage all modes of $y$ to be separated from each other. In conclusion, our final loss is:
\begin{equation}
\begin{split}
\mathcal{L} = & \mathcal{L}_{\vy}^{TrSm} + \alpha   \mathcal{L}_{\vx}^{Tr} + \beta\mathcal{L}_{c}
\label{eq:loss_triplet_softmax_and_triplet}
\end{split}
\end{equation}
where we found that $\alpha=\beta=1$ works reasonably well for all scenarios.


\head{Implementation details.} For all the three datasets, we train our model for $7$ epochs using 8 GPUs. We set learning rate as $0.001$ for the first $5$ epochs and $0.0001$ for the rest $2$ epochs. We initialize each branch with weights pre-trained on COCO \citet{lin2014microsoft}. For the word vectors, we used the \texttt{gensim} library \citet{rehurek_lrec} for both word2vec and node2vec\footnote{\url{https://github.com/aditya-grover/node2vec}} \citet{grover2016node2vec}. For the triplet loss, we set $m=0.2$ as the default value.

For the VRD and VG200 datasets, we need to predict whether a box pair has relationship, since unlike VG80k where we use ground-truth boxes, here we want to use general proposals that might contain non-relationships. In order for that, we add an additional ``unknown'' category to the relation categories. The word ``unknown'' is semantically dissimilar with any of the relations in these datasets, hence its word vector is far away from those relations' vectors.

There is a critical factor that significantly affects our triplet-softmax loss. Since we use cosine similarity, $s(\cdot,\cdot)$ is equivalent to dot product of two normalized vectors. We empirically found that simply feeding normalized vector could cause gradient vanishing problem, since gradients are divided by the norm of input vector when back-propagated. This is also observed in \citet{bell16ion} where it is necessary to scale up normalized vectors for successful learning. Similar with \citet{bell16ion}, we set the scalar to a value that is close to the mean norm of the input vectors and multiply $s(\cdot,\cdot)$ before feeding to the softmax layer. We set the scalar to $3.2$ for VG80k and $3.0$ for VRD in all experiments.

\head{ROI Sampling.} One of the critical things that powers Fast-RCNN is the well-designed ROI sampling during training. It ensures that for most ground-truth boxes, each has $32$ positive ROIs and $128-32=96$ negative ROIs, where positivity is defined as overlap IoU $>=0.5$. In our setting, ROI sampling is similar for the subject/object branch, while for the relation branch, positivity is defined as both subject and object IoUs $>=0.5$. Accordingly, we sample $64$ subject ROIs with $32$ unique positives and $32$ unique negatives, and do the same thing for object ROIs. Then we pair all the $64$ subject ROIs with $64$ object ROIs to get $4096$ ROI pairs as relationship candidates. For each candidate, if both ROIs' IoU $>=0.5$ we mark it as positive, otherwise negative. We finally sample $32$ positive and $96$ negative relation candidates and use the union of each ROI pair as a relation ROI. In this way we end up with a consistent number of positive and negative ROIs for the relation branch.

%% file: experiments.tex
\section{Experiments}
\label{sec:experiments}

\head{Datasets.} We present experiments on three datasets, the original \emph{Visual Genome} (VG80k) \citep{krishna2017visual}, the version of \emph{Visual Genome} with 200 categories (VG200) \citep{xu2017scenegraph}, and \emph{Visual Relationship Detection} (VRD) dataset \citep{lu2016visual}.

\begin{itemize}
\itemsep0em
\item \head{VRD.} The VRD dataset \citep{lu2016visual} contains 5,000 images with 100 object categories and 70 relations. In total, VRD contains 37,993 relation annotations with 6,672 unique relations and 24.25 relationships per object category. We follow the same train/test split as in \citet{lu2016visual} to get 4,000 training images and 1,000 test images. We use this dataset to demonstrate that our model can work reasonably well on small dataset with small category space, even though it is designed for large-scale settings.
\item \head{VG200.} We also train and evaluate our model on a subset of VG80k which is widely used in previous methods \citep{xu2017scenegraph,newell2017pixels,zellers2018scenegraphs,jianwei2018}. There are totally $150$ object categories and $50$ predicate categories in this dataset. We use the same train/test splits as in \citet{xu2017scenegraph}. Similarly with VRD, the purpose here is to show our model is also state-of-the-art in large-scale sample but small-scale category settings.
\item \head{VG80k.} We use the latest version of Visual Genome (VG v1.4) \citep{krishna2017visual} that contains $108,077$ images with $21$ relationships on average per image. We follow \citet{densecap} and split the data into $103,077$ training images and $5,000$ testing images. Since text annotations of VG are noisy, we first clean it by removing non-alphabet characters and stop words, and use the \texttt{autocorrect} library to correct spelling. Following that, we check if all words in an annotation exist in the word2vec dictionary \citep{mikolov2013distributed} and remove those that do not. We run this cleaning process on both training and testing set and get $99,961$ training images and $4,871$ testing images, with $53,304$ object categories and $29,086$ relation categories. We further split the training set into $97,961$ training and $2,000$ validation images.\footnote{We will release the cleaned annotations along with our code.}
\end{itemize}

\head{Evaluation protocol.}
For VRD, we use the same evaluation metrics used in \citet{yu17iccv}, which runs relationship detection using non-ground-truth proposals and reports recall rates using the top 50 and 100 relationship predictions, with $k=1,10,70$ relations per relationship proposal before taking the top 50 and 100 predictions.

For VG200, we use the same evaluation metrics used in \citet{zellers2018scenegraphs}, which uses three modes: 1) \textbf{predicate classification}: predict predicate labels given ground truth subject and object boxes and labels; 2) \textbf{scene graph classification}: predict subject, object and predicate labels given ground truth subject and object boxes; 3) \textbf{scene graph detection}: predict all the three labels and two boxes. Recalls under the top 20, 50, 100 predictions are used as metrics. The mean is computed over the 3 evaluation modes over R@50 and R@100 as in \citet{zellers2018scenegraphs}.

For VG80k, we evaluate all methods on the whole $53,304$ object and $29,086$ relation categories. We use ground-truth boxes as relationship proposals, meaning there is no localization errors and the results directly reflect recognition ability of a model. We use the following metrics to measure performance:
(1) top1, top5, and top10 accuracy,
(2) mean reciprocal ranking (rr), defined as $\frac{1}{M} \sum_{i=1}^{M} \frac{1}{rank_i}$,
(3) mean ranking (mr), defined as $\frac{1}{M} \sum_{i=1}^{M} {rank_i}$, smaller is better.

\begin{table*}[t]
\begin{adjustbox}{max width=1\textwidth,center}
\begin{tabular}{l c c c c | c c c c c c | c c c c c c}
\hline
& \multicolumn{2}{c}{Relationship} & \multicolumn{2}{c}{Phrase} & \multicolumn{6}{c}{Relationship Detection} & \multicolumn{6}{c}{Phrase Detection} \\
& \multicolumn{4}{c}{free k} & \multicolumn{2}{c}{k = 1} & \multicolumn{2}{c}{k = 10}  & \multicolumn{2}{c}{k = 70}  & \multicolumn{2}{c}{k = 1} & \multicolumn{2}{c}{k = 10}  & \multicolumn{2}{c}{k = 70} \\
Recall at & 50 & 100 & 50 & 100 & 50 & 100 & 50 & 100 & 50 & 100 & 50 & 100 & 50 & 100 & 50 & 100 \\
\hline
\multicolumn{4}{l}{\rule{0pt}{15pt}\textbf{w/ proposals from \citep{lu2016visual}}} &  &  &  &  & &  &  &  &  &  &  &  \\
CAI*\citep{Zhuang_2017_ICCV} & 15.63 & 17.39 & 17.60 & 19.24 & - & - & - & - & - & - & - & - & - & - & - & - \\
Language cues\citep{plummerPLCLC2017} & 16.89 & 20.70 & 15.08 & 18.37 & - & - & 16.89 & 20.70 & - & - & - & - & 15.08 & 18.37 & - & - \\
VRD\citep{lu2016visual} & 17.43 & 22.03 & 20.42 & 25.52 & 13.80 & 14.70 & 17.43 & 22.03 & 17.35 & 21.51 & 16.17 & 17.03 & 20.42 & 25.52 & 20.04 & 24.90 \\
Ours & \textbf{19.18} & \textbf{22.64} & \textbf{21.69} & \textbf{25.92} & \textbf{16.08} & \textbf{17.07} & \textbf{19.18} & \textbf{22.64} & \textbf{18.89} & \textbf{22.35} & \textbf{18.32} & \textbf{19.78} & \textbf{21.69} & \textbf{25.92} & \textbf{21.39} & \textbf{25.65} \\ 
\multicolumn{4}{l}{\rule{0pt}{15pt}\textbf{w/ better proposals}} &  &  &  &  & &  &  &  &  &  &  &  \\
DR-Net*\citep{dai2017detecting} & 17.73 & 20.88 & 19.93 & 23.45 \rule{3pt}{0pt}& - & - & - & - & - & - & - & - & - & - & - & - \\
ViP-CNN\citep{LiCVPR2017} & 17.32 & 20.01 &  22.78 & 27.91 & 17.32 & 20.01 & - & - & - & - & 22.78 & 27.91 & - & - & - & - \\
VRL\citep{liang2017deep} & 18.19 & 20.79 & 21.37 & 22.60  & 18.19 & 20.79 & - & - & - & - & 21.37 & 22.60 & - & - & - & - \\
PPRFCN*\citep{zhang2017ppr} & 14.41 & 15.72 & 19.62 & 23.75  & - & - & - & - & - & - & - & - & - & - & - & - \\
VTransE* & 14.07 & 15.20 & 19.42 & 22.42  & - & - & - & - & - & - & - & - & - & - & - & - \\
SA-Full*\citep{Peyre17} & 15.80 & 17.10 & 17.90 & 19.50  & - & - & - & - & - & - & - & - & - & - & - & - \\
CAI*\citep{Zhuang_2017_ICCV} & 20.14 & 23.39 & 23.88 & 25.26 & - & - & - & - & - & - & - & - & - & - & - & - \\
KL distilation\citep{yu17iccv} & 22.68 & 31.89 & 26.47 & 29.76 & 19.17 & 21.34 & 22.56 & 29.89 & 22.68 & 31.89 & 23.14 & 24.03 & 26.47 & 29.76 & 26.32 & 29.43 \\
Zoom-Net\citep{Yin_2018_ECCV} & 21.37 & 27.30 & 29.05 & 37.34 & 18.92 & 21.41 & - & - & 21.37 & 27.30 & 24.82 & 28.09 & - & - & 29.05 & 37.34 \\
CAI + SCA-M\citep{Yin_2018_ECCV} & 22.34 & 28.52 & 29.64 & 38.39 & 19.54 & 22.39 & - & - & 22.34 & 28.52 & 25.21 & 28.89 & - & - & 29.64 & 38.39 \\
Ours & \textbf{26.98} & \textbf{32.63} & \textbf{32.90} & \textbf{39.66} & \textbf{23.68} & \textbf{26.67} & \textbf{26.98} & \textbf{32.63} & \textbf{26.98} & \textbf{32.59} & \textbf{28.93} & \textbf{32.85} & \textbf{32.90} & \textbf{39.66} & \textbf{32.90} & \textbf{39.64} \\
\hline
\end{tabular}
\end{adjustbox}
\captionsetup{font=small}
\caption{Comparison with state-of-the-art on the VRD dataset.}
\label{tab:vrd}
\end{table*}

\begin{table*}[t!]
\centering
\begin{adjustbox}{max width=1\textwidth,center}
\begin{tabular}{l c c c | c c c | c c c}
\hline
& \multicolumn{3}{c}{Scene Graph Detection} & \multicolumn{3}{c}{Scene Graph Classification} & \multicolumn{3}{c}{Predicate Classification} \\
Recall at & 20 & 50 & 100 & 20 & 50 & 100 & 20 & 50 & 100 \\
\hline
VRD\citep{lu2016visual} & - & 0.3 & 0.5 & - & 11.8 & 14.1 & - & 27.9 & 35.0 \\
Message Passing\citep{xu2017scenegraph} & - & 3.4 & 4.2 & - & 21.7 & 24.4 & - & 44.8 & 53.0 \\
Message Passing+ & 14.6 & 20.7 & 24.5 & 31.7 & 34.6 & 35.4 & 52.7 & 59.3 & 61.3 \\
Associative Embedding\citep{newell2017pixels} & 6.5 & 8.1 & 8.2 & 18.2 & 21.8 & 22.6 & 47.9 & 54.1 & 55.4 \\
Frequency & 17.7 & 23.5 & 27.6 & 27.7 & 32.4 & 34.0 & 49.4 & 59.9 & 64.1 \\
Frequency+Overlap & 20.1 & 26.2 & 30.1 & 29.3 & 32.3 & 32.9 & 53.6 & 60.6 & 62.2 \\
MotifNet-LeftRight \citep{zellers2018scenegraphs}& \textbf{21.4} & 27.2 & 30.3 & 32.9 & 35.8 & 36.5 & 58.5 & 65.2 & 67.1 \\
Ours & 20.7 & \textbf{27.9} & \textbf{32.5} & \textbf{36.0} & \textbf{36.7} & \textbf{36.7} & \textbf{66.8} & \textbf{68.4} & \textbf{68.4} \\
\hline
\end{tabular}
\end{adjustbox}
\captionsetup{font=small}
\caption{Comparison with state-of-the-art on the VG200 dataset.}
\label{tab:vg200}
\end{table*}

\subsection{Evaluation of Relationship Detection on VRD}
\label{subsec:vrd}

We first validate our model on VRD dataset with comparison to state-of-the-art methods using the metrics presented in \citet{yu17iccv} in Table \ref{tab:vrd}. Note that there is a variable $k$ in this metric which is the number of relation candidates when selecting top50/100. Since not all previous methods specified $k$ in their evaluation, we first report performance in the ``free $k$'' column when considering $k$ as a hyper-parameter that can be cross-validated. For methods where the $k$ is reported for 1 or more values, the column reports the performance using the best $k$. We then list all available results with specific $k$ in the right two columns.


For fairness, we split the table in two parts. The top part lists methods that use the same proposals from \citet{lu2016visual}, while the bottom part lists methods that are based on a different set of proposals, and ours uses better proposals obtained from Faster-RCNN as previous works.
We can see that we outperform all other methods with proposals from \citet{lu2016visual} even without using message-passing-like post processing as in \citet{LiCVPR2017,dai2017detecting}, and also very competitive to the overall best performing method from \citet{yu17iccv}. Note that although spatial features could be advantageous for VRD according to previous methods, we do not use them in our model in concern of large-scale settings. We expect better performance if integrating spatial features for VRD, but for model consistency we do experiments without it everywhere.
\subsection{Scene Graph Classification \& Detection on VG200}
\label{subsec:vg200}

We present our results in Table \ref{tab:vg200}.
Note that scene graph classification isolates the factor of subject/object localization accuracy by using ground truth subject/object boxes, meaning that it focuses more on the relationship recognition ability of a model, and predicate classification focuses even more on it by using ground truth subject/object boxes and labels. It is clear that the gaps between our model and others are higher on scene graph/predicate classification, meaning our model displays superior relation recognition ability.

\begin{table*}[t!]
\centering
\begin{adjustbox}{max width=0.9\textwidth,center}
\begin{tabular}{l c c c c c | c c c c c}
\hline
& \multicolumn{5}{c}{Relationship Triplet} & \multicolumn{5}{c}{Relation} \\
 & top1 & top5 & top10 & rr & mr & top1 & top5 & top10 & rr & mr \\
\hline
\textbf{All classes} & & & & & & & & & & \\
3-branch Fast-RCNN & 9.73 & 41.95 & 55.19 & 52.10 & 16.36 & 36.00 & 69.59 & 79.83 & 50.77 & 7.81 \\
ours w/ triplet & 8.01 & 27.06 & 35.27 & 40.33 & 32.10 & 37.98 & 61.34 & 69.60 & 48.28 & 14.12 \\
ours w/ softmax & 14.53 & 46.33 & 57.30 & 55.61 & 16.94 & 49.83 & 76.06 & 82.20 & 61.60 & 8.21 \\
ours final & \textbf{15.72} & \textbf{48.83} & \textbf{59.87} & \textbf{57.53} & \textbf{15.08} & \textbf{52.00} & \textbf{79.37} & \textbf{85.60} & \textbf{64.12} & \textbf{6.21} \\
\textbf{Tail classes} & & & & & & & & & & \\
3-branch Fast-RCNN & 0.32 & 3.24 & 7.69 & 24.56 & 49.12 & 0.91 & 4.36 & 9.77 & 4.09 & 52.19 \\
ours w/ triplet & 0.02 & 0.29 & 0.58 & 7.73 & 83.75 & 0.12 & 0.61 & 1.10 & 0.68 & 86.60 \\
ours w/ softmax & 0.00 & 0.07 & 0.47 & 20.36 & 58.50 & 0.00 & 0.08 & 0.55 & 1.11 & 65.02 \\
ours final & \textbf{0.48} & \textbf{13.33} & \textbf{28.12} & \textbf{43.26} & \textbf{45.48} & \textbf{0.96} & \textbf{7.61} & \textbf{16.36} & \textbf{5.56} & \textbf{45.70} \\
\hline
\end{tabular}
\end{adjustbox}
\captionsetup{font=small}
\caption{Results on all relation classes and tail classes ($\#occurrence \leq 1024$) in VG80k. Note that since VG80k is extremely imbalanced, classes with no greater than 1024 occurrences are still in the tail. In fact, there are more than 99\% of relation classes but only 10.04\% instances of these classes that occur for no more than 1024 times.}
\label{tab:comparison}
\end{table*}

\begin{figure*}[t!]

  \centering
  \begin{subfigure}{0.99\columnwidth}
    \centering
    \begin{subfigure}{\columnwidth}
      \centering
      \includegraphics[width=\textwidth]{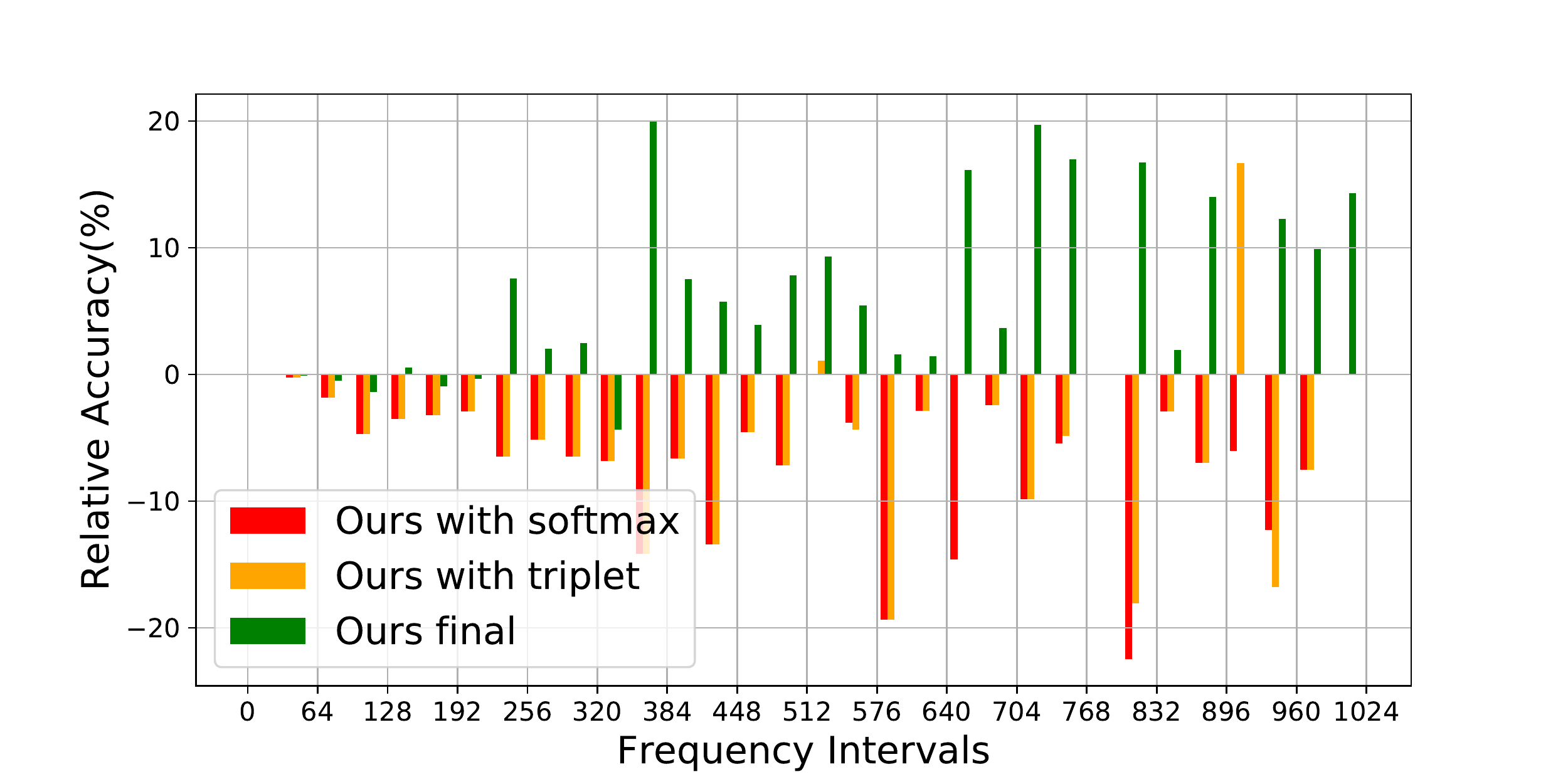}
    \end{subfigure}
  \captionsetup{font=small}
  \caption{Top 5 rel triplet}
  \end{subfigure}
  \centering
  \begin{subfigure}{0.99\columnwidth}
    \centering
    \begin{subfigure}{\columnwidth}
      \centering
      \includegraphics[width=\textwidth]{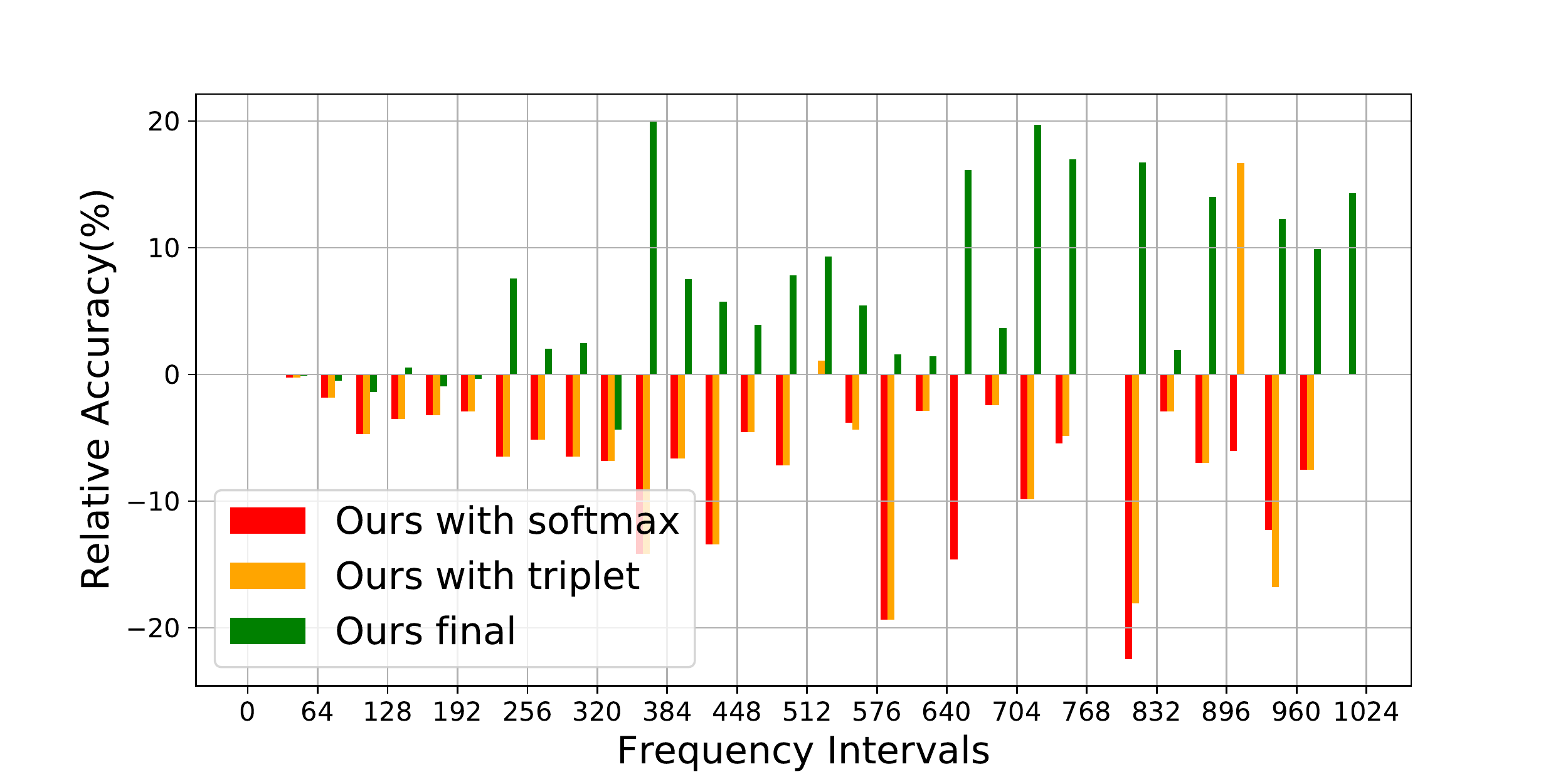}
    \end{subfigure}
  \captionsetup{font=small}
  \caption{Top 5 relation}
  \end{subfigure}
\setlength\belowcaptionskip{-2ex}
\captionsetup{font=small}
\caption{Top-5 relative accuracies against the 3-branch Fast-RCNN baseline in the tail intervals. The intervals are defined as bins of 32 from 1 to 1024 occurrences of the relation classes.}
\label{fig:interval_tail_baselines}
\end{figure*}

\subsection{Relationship Recognition on VG80k}
\label{subsec:vg}

\head{Baselines.} Since there is no previous method that has been evaluated in our large-scale setting, we carefully design 3 baselines to compare with. 1) 3-branch Fast-RCNN: an intuitively straightforward model is a Fast-RCNN with a shared $conv1$ to $conv5$ backbone and 3 $fc$ branches for subject, relation and object respectively, where the subject and object branches share weights since they are essentially an object detector; 2) our model with softmax loss: we replace our loss with softmax loss; 3) our model with triplet loss: we replace our loss with triplet loss.

\head{Results.} As shown in Table \ref{tab:comparison}, we can see that our loss is the best for the general case where all instances from all classes are considered. The baseline has reasonable performance but is clearly worse than ours with softmax, demonstrating that our visual module is critical for efficient learning. Ours with triplet is worse than ours with softmax in the general case since triplet loss is not discriminative enough among the massive data. However it is the opposite for tail classes (i.e., $\#occurrence \leq 1024$), since recognition of infrequent classes can benefit from the transferred knowledge learned from frequent classes, which the softmax-based model is not capable of. Another observation is that although the 3-branch Fast-RCNN baseline works poorly in the general case, it is better than our model with softmax. Since the main difference of them is with and without visual feature concatenation, it means that integrating subject and object features does not necessarily helps infrequent relation classes. This is because subject and object features could lead to strong prior on the relation, resulting in lower chance of predicting infrequent relation when using softmax. For example, when seeing a rare image where the relationship is ``dog ride horse'', subject being ``dog'' and object being ``horse'' would give very little probability to the relation ``ride'', even though it is the correct answer. Our model alleviates this problem by not mapping visual features directly to the discrete categorical space, but to a continuous embedding space where visual similarity is preserved. Therefore, when seeing the visual features of ``dog'', ``horse'' and the whole ``dog ride horse'' context, our model is able to associate them with a visually similar relationship ``person ride horse'' and correctly output the relation ``ride''.

\subsection{Ablation Study}
\label{subsec:ablation}

\begin{table*}[t!]
\centering
\begin{adjustbox}{max width=0.8\textwidth,center}
\begin{tabular}{l c c c c c | c c c c c}
\hline
& \multicolumn{5}{c}{\textbf{Relationship Triplet}} & \multicolumn{5}{c}{\textbf{Relation}} \\
Methods & top1 & top5 & top10 & rr & mr & top1 & top5 & top10 & rr & mr \\
\hline
wiki & 15.59 & 46.03 & 54.78 & 52.45 & 25.31 & 51.96 & 78.56 & 84.38 & 63.61 & 8.61 \\
relco & 15.58 & 46.63 & 55.91 & 54.03 & 22.23 & 52.00 & 79.06 & 84.75 & 63.90 & 7.74 \\
wiki + relco & \textbf{15.72} & \textbf{48.83} & \textbf{59.87} & \textbf{57.53} & \textbf{15.08} & \textbf{52.00} & \textbf{79.37} & \textbf{85.60} & \textbf{64.12} & \textbf{6.21} \\
wiki + node2vec & 15.62 & 47.58 & 57.48 & 54.75 & 20.93 & 51.92 & 78.83 & 85.01 & 63.86 & 7.64 \\
\hline
0 sem layer & 11.21 & 28.78 & 34.84 & 38.64 & 43.49 & 44.66 & 60.06 & 64.74 & 51.60 & 24.74 \\
1 sem layer & \textbf{15.75} & 48.23 & 58.28 & 55.70 & 19.15 & 51.82 & 78.94 & 85.00 & 63.79 & 7.63 \\
2 sem layer & 15.72 & \textbf{48.83} & \textbf{59.87} & \textbf{57.53} & \textbf{15.08} & \textbf{52.00} & \textbf{79.37} & \textbf{85.60} & \textbf{64.12} & \textbf{6.21} \\
3 sem layer & 15.49 & 48.42 & 58.75 & 56.98 & 15.83 & \textbf{52.00} & 79.19 & 85.08 & 63.99 & 6.40 \\
\hline
no concat & 10.47 & 42.51 & 54.51 & 51.51 & 20.16 & 36.96 & 70.44 & 80.01 & 51.62 & 9.26 \\
early concat & 15.09 & 45.88 & 55.72 & 54.72 & 19.69 & 49.54 & 75.56 & 81.49 & 61.25 & 8.82 \\
late concat & 15.57 & 47.72 & 58.05 & 55.34 & 19.27 & 51.06 & 78.15 & 84.47 & 63.03 & 7.90 \\
both concat & \textbf{15.72} & \textbf{48.83} & \textbf{59.87} & \textbf{57.53} & \textbf{20.62} & \textbf{52.00} & \textbf{79.37} & \textbf{85.60} & \textbf{64.12} & \textbf{6.21} \\
\hline
$\mathcal{L}_{y}$ & 15.21 & 47.28 & 57.77 & 55.06 & 19.12 & 50.67 & 78.21 & 84.70 & 62.82 & 7.31 \\
$\mathcal{L}_{y}$ + $\mathcal{L}_{x}$ & 15.07 & 47.37 & 57.85 & 54.92 & 19.59 & 50.60 & 78.06 & 84.40 & 62.71 & 7.60 \\
$\mathcal{L}_{y}$ + $\mathcal{L}_{c}$ & 15.53 & 47.97 & 58.49 & 55.78 & 18.55 & 51.48 & 78.99 & 84.90 & 63.59 & 7.32 \\
$\mathcal{L}_{y}$ + $\mathcal{L}_{x}$ + $\mathcal{L}_{c}$ & \textbf{15.72} & \textbf{48.83} & \textbf{59.87} & \textbf{57.53} & \textbf{15.08} & \textbf{52.00} & \textbf{79.37} & \textbf{85.60} & \textbf{64.12} & \textbf{6.21} \\
\hline
\end{tabular}
\end{adjustbox}
\setlength\belowcaptionskip{-2ex}
\captionsetup{font=small}
\caption{Ablation study of our model on VG80k.
}
\label{tab:ablation_study}
\end{table*}

\begin{table}[t!]
\centering
\begin{adjustbox}{max width=1\linewidth,center}
\begin{tabular}{c | c c c c c | c c c c c}
\hline
& \multicolumn{5}{c}{\textbf{Relationship Triplet}} & \multicolumn{5}{c}{\textbf{Relation}} \\
$\lambda$ = & top1 & top5 & top10 & rr & mr & top1 & top5 & top10 & rr & mr \\
\hline
1.0 & 0.00 & 0.61 & 3.77 & 22.43 & 48.24 & 0.04 & 1.12 & 5.97 & 4.11 & 21.39 \\
2.0 & 8.48 & 27.63 & 34.26 & 35.25 & 46.28 & 44.94 & 70.60 & 76.63 & 56.69 & 13.20 \\
3.0 & 14.19 & 39.22 & 46.71 & 48.80 & 29.65 & 51.07 & 74.61 & 78.74 & 61.74 & 10.88 \\
4.0 & \textbf{15.72} & 47.19 & 56.94 & 54.80 & 20.85 & 51.67 & 78.66 & 84.23 & 63.53 & 8.68 \\
5.0 & \textbf{15.72} & \textbf{48.83} & \textbf{59.87} & \textbf{57.53} & \textbf{15.08} & \textbf{52.00} & \textbf{79.37} & \textbf{85.60} & \textbf{64.12} & \textbf{6.21} \\
6.0 & 15.32 & 47.99 & 58.10 & 55.57 & 18.67 & 51.60 & 78.95 & 85.05 & 63.62 & 7.23 \\
7.0 & 15.11 & 44.72 & 54.68 & 54.04 & 20.82 & 51.23 & 77.37 & 83.37 & 62.95 & 7.86 \\
8.0 & 14.84 & 45.12 & 54.95 & 54.07 & 20.56 & 51.25 & 77.67 & 83.36 & 62.97 & 7.81 \\
9.0 & 14.81 & 45.72 & 55.81 & 54.29 & 20.10 & 50.88 & 78.59 & 84.70 & 63.08 & 7.21 \\
10.0 & 14.71 & 45.62 & 55.71 & 54.19 & 20.19 & 51.07 & 78.64 & 84.78 & 63.21 & 7.26 \\
\hline
\end{tabular}
\end{adjustbox}
\setlength\belowcaptionskip{-2ex}
\captionsetup{font=small}
\caption{Performances of our model on VG80k validation set with different values of the scaling factor.
We use scaling factor $\lambda=5.0$ for all our experiments on VG80k.
}
\label{tab:ours_scaling}
\end{table}


\head{Variants of our model.} We explore variants of our model in 4 dimensions: 1) the semantic embeddings fed to the semantic module; 2) structure of the semantic module; 3) structure of the visual module; 4) the losses. The default settings of them are 1) using \textit{wiki + relco}; 2) 2 semantic layer; 3) with both visual concatenation; 4) with all the 3 loss terms. We fix the other 3 dimensions as the default settings when exploring one of them.

\head{The scaling factor before softmax.} As mentioned in the implementation details, this value scales up the output by a value that is close to the average norm of the input and prevents gradient vanishing caused by the normalization. Specifically, for Eq(7) in the paper we use $s(\vx,\vy)=\lambda\frac{\vx^T\vy}{||\vx||||\vy||}$ where $\lambda$ is the scaling factor. In Table \ref{tab:ours_scaling} we show results of our model when changing the value of the scaling factor applied before the softmax layer. We observe that when the value is close to the average norm of all input vectors (i.e., 5.0), we achieve optimal performance, although slight difference of this value does not change results too much (i.e., when it is 4.0 or 6.0). It is clear that when the scaling factor is 1.0, which is equivalent to training without scaling, the model is not sufficiently trained. We therefore pick 5.0 for this scaling factor for all the other experiments on VG80k.

\head{Which semantic embedding to use?} We explore 4 settings: 1) \textit{wiki} and 2) \textit{relco} use wikipedia and relationship-level co-occurrence embedding alone, while 3) \textit{wiki + relco} and 4) \textit{wiki + node2vec} use concatenation of two embeddings. The intuition of concatenating \textit{wiki} with \textit{relco} and \textit{node2vec} is that \textit{wiki} contains common knowledge acquired outside of the dataset, while \textit{relco} and \textit{node2vec} are trained specifically on VG80k, and their combination provides abundant information for the semantic module. As shown in Table \ref{tab:ablation_study}, fusion of \textit{wiki} and \textit{relco} outperforms each one alone with clear margins. We found that using \textit{node2vec} alone does not perform reasonably, but \textit{wiki + node2vec} is competitive to others, demonstrating the efficacy of concatenation.

\head{Number of semantic layers.} We also study how many, if any, layers are necessary to embed the word vectors. As it is shown in Table \ref{tab:ablation_study}, directly using the word vectors (0 semantic layers) is not a good substitute of our learned embedding; raw word vectors are learned to represent as much associations between words as possible, but not to distinguish them. We find that either 1 or 2 layers give similarly good results and 2 layers are slightly better, though performance starts to degrade when adding more layers.

\head{Are both visual feature concatenations necessary?}
In Table \ref{tab:ablation_study}, ``early concat'' means using only the first concatenation of the three branches, and ``late concat'' means the second. Both early and late concatenation boost performance significantly compared to no concatenation, and it is the best with both. Another observation is that late concatenation is better than early alone. We believe the reason is, as mentioned above, relations are naturally conditioned on and constrained by subjects and objects, e.g., given ``man'' as subject and ``chair'' as object, it is highly likely that the relation is ``sit on''. Since late concatenation is at a higher level, it integrates features that are more semantically close to the subject and object labels, which gives stronger prior to the relation branch and affects relation prediction more than the early concatenation.

\head{Do all the losses help?}
In order to understand how each loss helps training, we trained 3 models of which each excludes one or two loss terms. We can see that using $\mathcal{L}_{y} + \mathcal{L}_{x}$ is similar with $\mathcal{L}_{y}$, and it is the best with all the three losses. This is because $\mathcal{L}_{x}$ pulls positive $x$ pairs close while pushes negative $x$ away. However, since $(x, y)$ is a many-to-one mapping (i.e., multiple visual features could have the same label), there is no guarantee that all $x$ with the same $y$ would be embedded closely, if not using $\mathcal{L}_{c}$. By introducing $\mathcal{L}_{c}$, $x$ with the same $y$ are forced to be close to each other, and thus the structural consistency of visual features is preserved.

\begin{table}[t!]
\centering
\begin{adjustbox}{max width=0.985\linewidth,center}
\begin{tabular}{c | c c c c c | c c c c c}
\hline
& \multicolumn{5}{c}{\textbf{Relationship Triplet}} & \multicolumn{5}{c}{\textbf{Relation}} \\
m = & top1 & top5 & top10 & rr & mr & top1 & top5 & top10 & rr & mr \\
\hline
0.1 & 7.77 & 29.84 & \textbf{38.53} & \textbf{42.29} & \textbf{28.13} & 36.50 & \textbf{63.50} & \textbf{70.20} & 47.48 & 14.20 \\
0.2 & \textbf{8.01} & \textbf{27.06} & 35.27 & 40.33 & 32.10 & \textbf{37.98} & 61.34 & 69.60 & \textbf{48.28} & \textbf{14.12} \\
0.3 & 5.78 & 24.39 & 33.26 & 37.03 & 34.55 & 36.75 & 58.65 & 64.86 & 46.62 & 20.62 \\
0.4 & 3.82 & 22.55 & 31.70 & 34.10 & 36.26 & 34.89 & 57.25 & 63.74 & 45.04 & 21.89 \\
0.5 & 3.14 & 19.69 & 30.01 & 31.63 & 38.25 & 33.65 & 56.16 & 62.77 & 43.88 & 23.19 \\
0.6 & 2.64 & 15.68 & 27.65 & 29.74 & 39.70 & 32.15 & 55.08 & 61.68 & 42.52 & 24.25 \\
0.7 & 2.17 & 11.35 & 24.55 & 28.06 & 41.47 & 30.36 & 54.20 & 60.60 & 41.02 & 25.23 \\
0.8 & 1.87 & 8.71 & 16.30 & 26.43 & 43.18 & 29.78 & 53.43 & 60.01 & 40.29 & 26.19 \\
0.9 & 1.43 & 7.44 & 11.50 & 24.76 & 44.83 & 28.35 & 51.73 & 58.74 & 38.89 & 27.27 \\
1.0 & 1.10 & 6.97 & 10.51 & 23.57 & 46.60 & 27.49 & 50.72 & 58.10 & 37.97 & 28.13 \\
\hline
\end{tabular}
\end{adjustbox}
\setlength\belowcaptionskip{-2ex}
\captionsetup{font=small}
\caption{Performances of triplet loss on VG80k validation set with different values of margin m.
We use margin $m=0.2$ for all our experiments in the main paper.
}
\label{tab:triplet_m}
\end{table}

\begin{figure*}[t!]
  \centering
  \begin{subfigure}{0.3\linewidth}
    \centering
    \begin{subfigure}{\linewidth}
      \centering
	  \includegraphics[width=\textwidth]{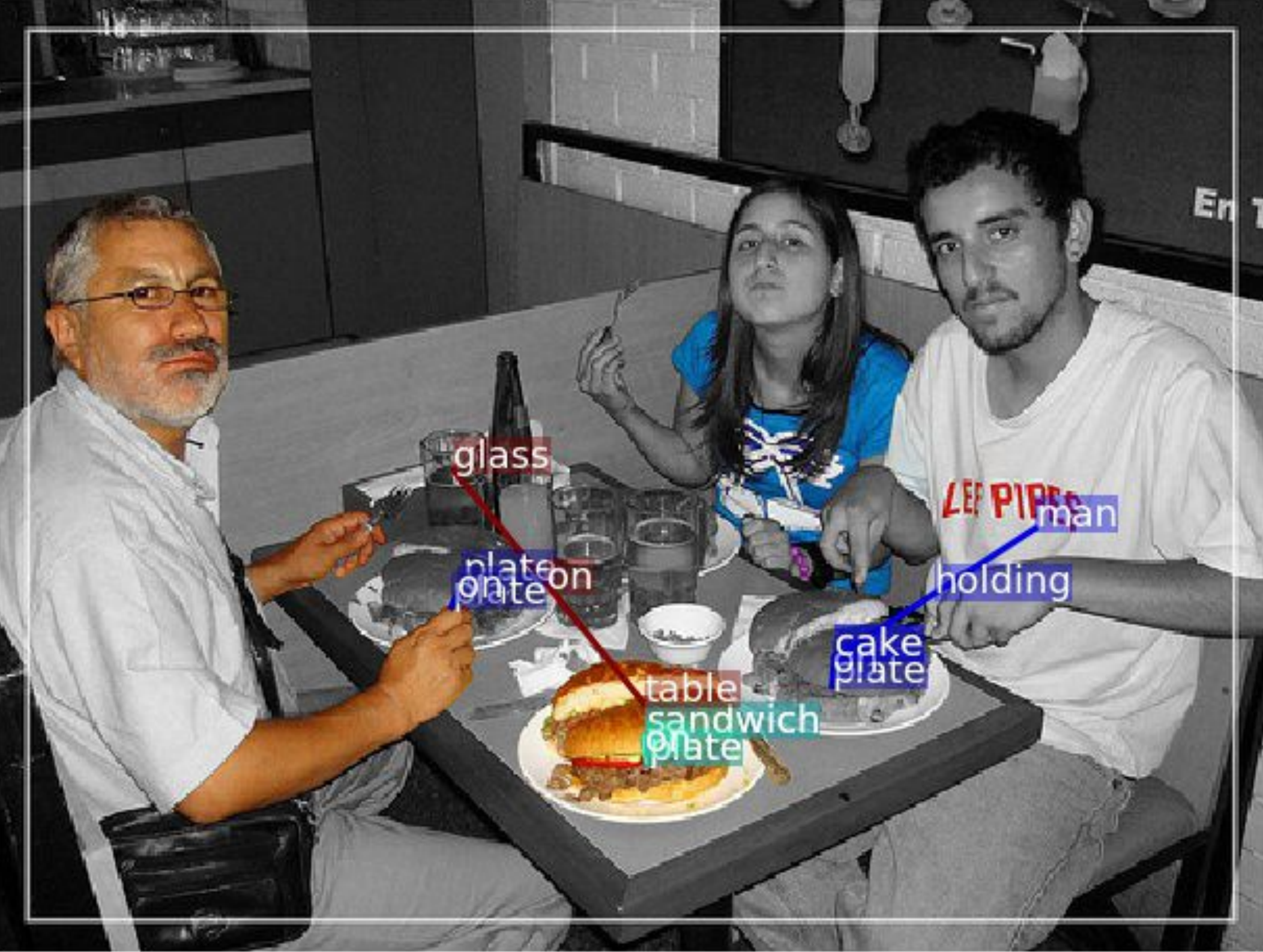}
    \end{subfigure}
  \end{subfigure}
  \centering
  \begin{subfigure}{0.335\linewidth}
    \centering
    \begin{subfigure}{\linewidth}
      \centering
      \includegraphics[width=\textwidth]{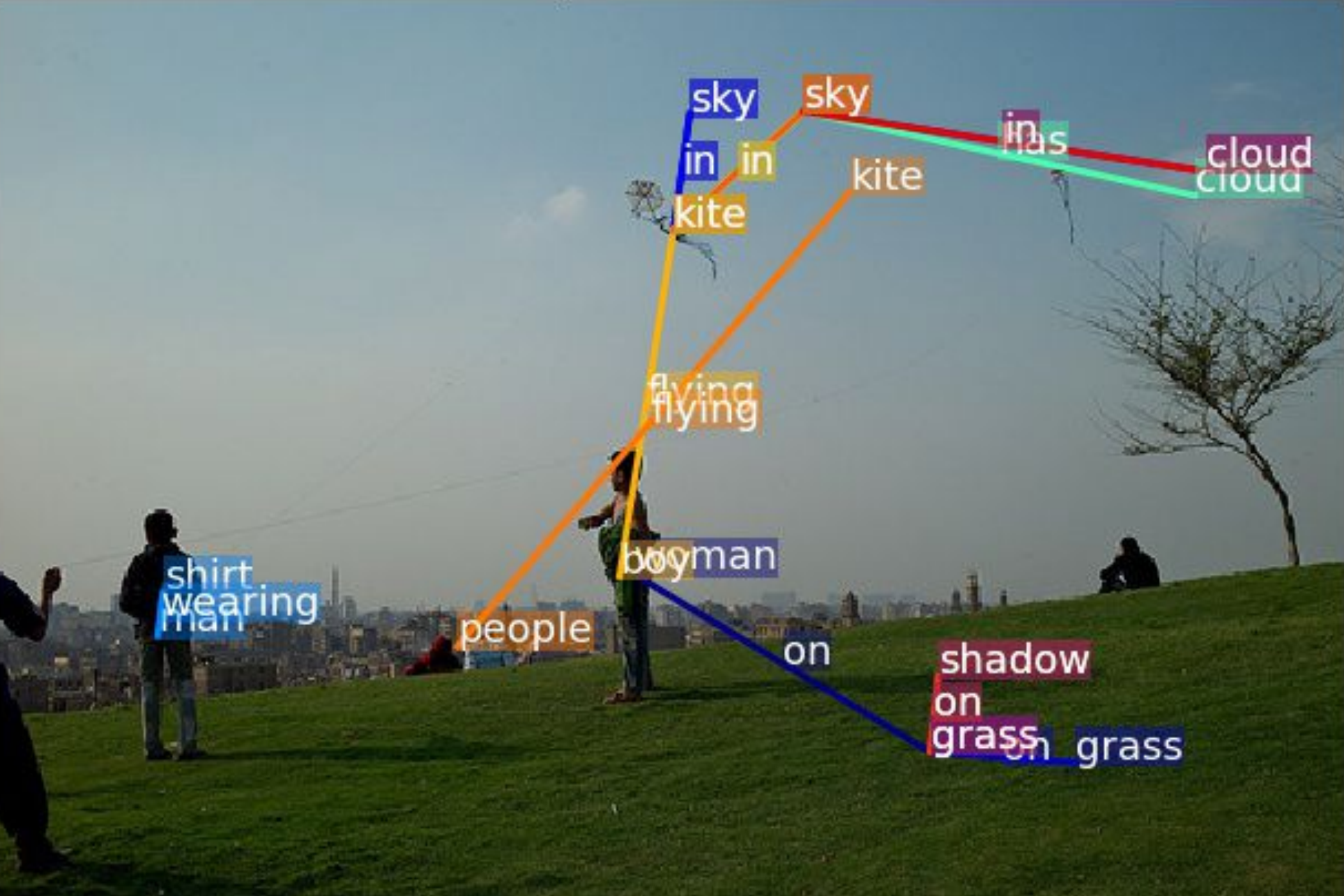}
    \end{subfigure}
  \end{subfigure}
  \begin{subfigure}{0.336\linewidth}
    \centering
    \begin{subfigure}{\linewidth}
      \centering
      \includegraphics[width=\textwidth]{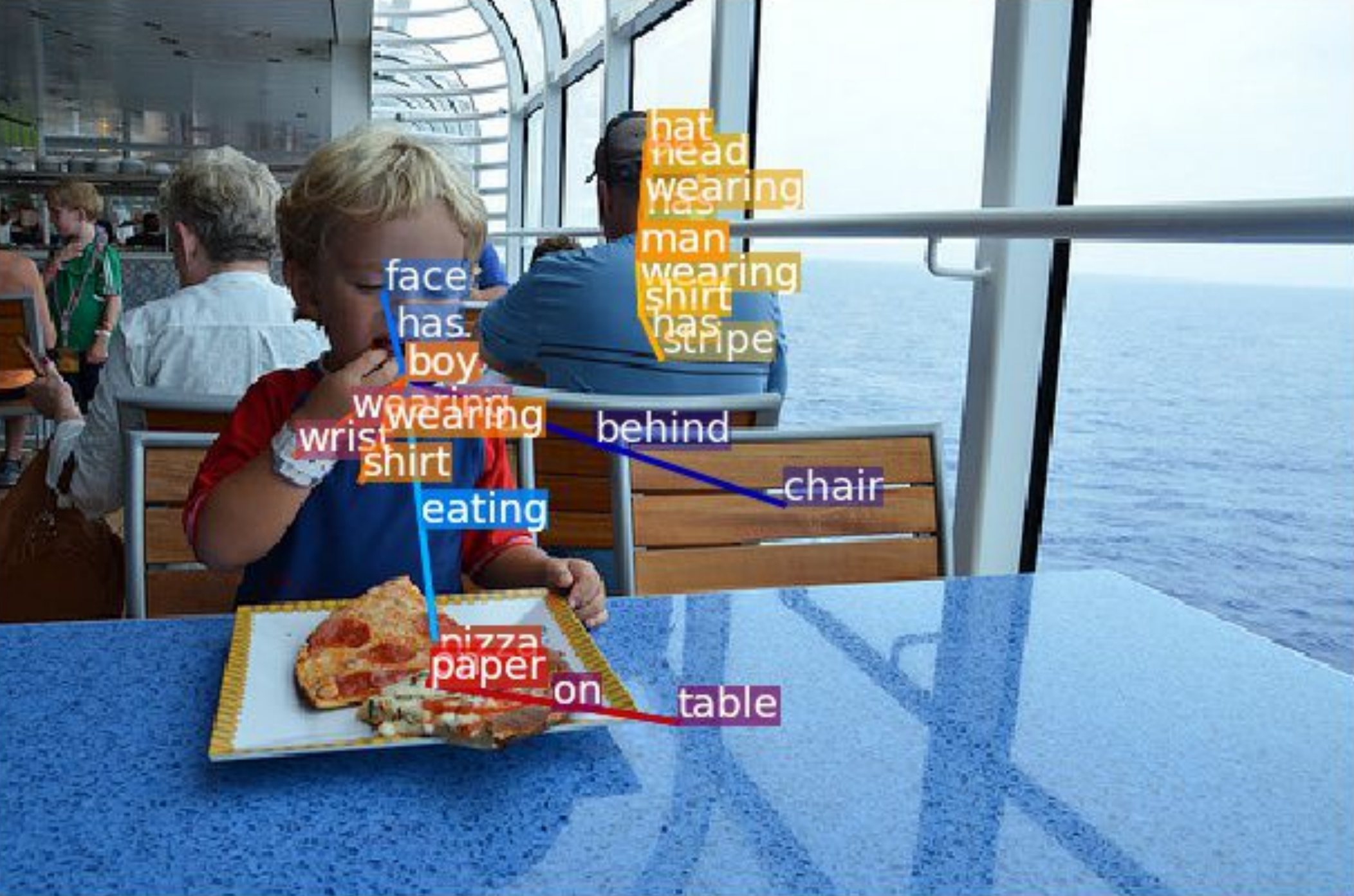}
    \end{subfigure}
  \end{subfigure}
  
  \centering
  \begin{subfigure}{0.285\linewidth}
    \centering
    \begin{subfigure}{\linewidth}
      \centering
	  \includegraphics[width=\textwidth]{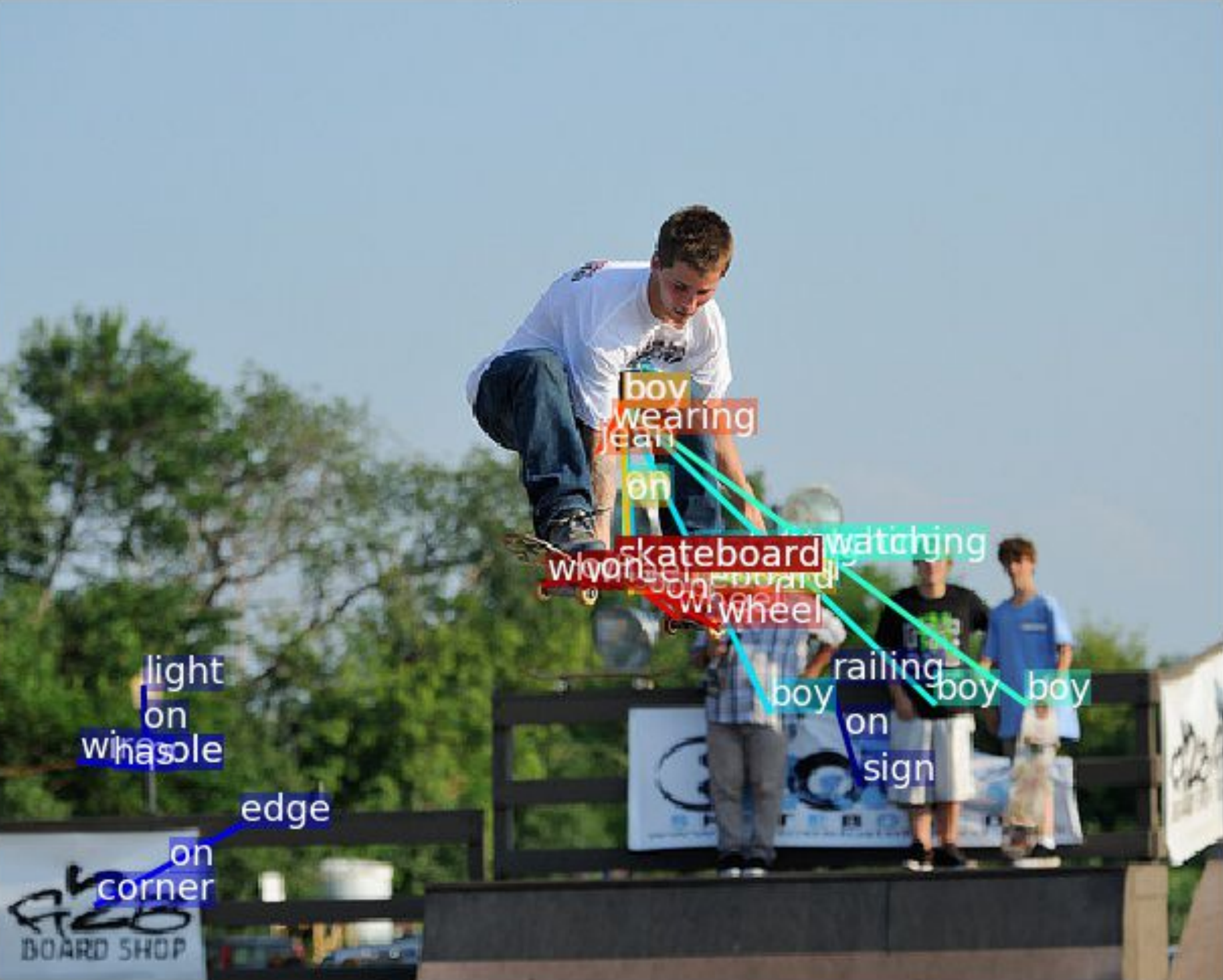}
    \end{subfigure}
  \end{subfigure}
  \centering
  \begin{subfigure}{0.342\linewidth}
    \centering
    \begin{subfigure}{\linewidth}
      \centering
      \includegraphics[width=\textwidth]{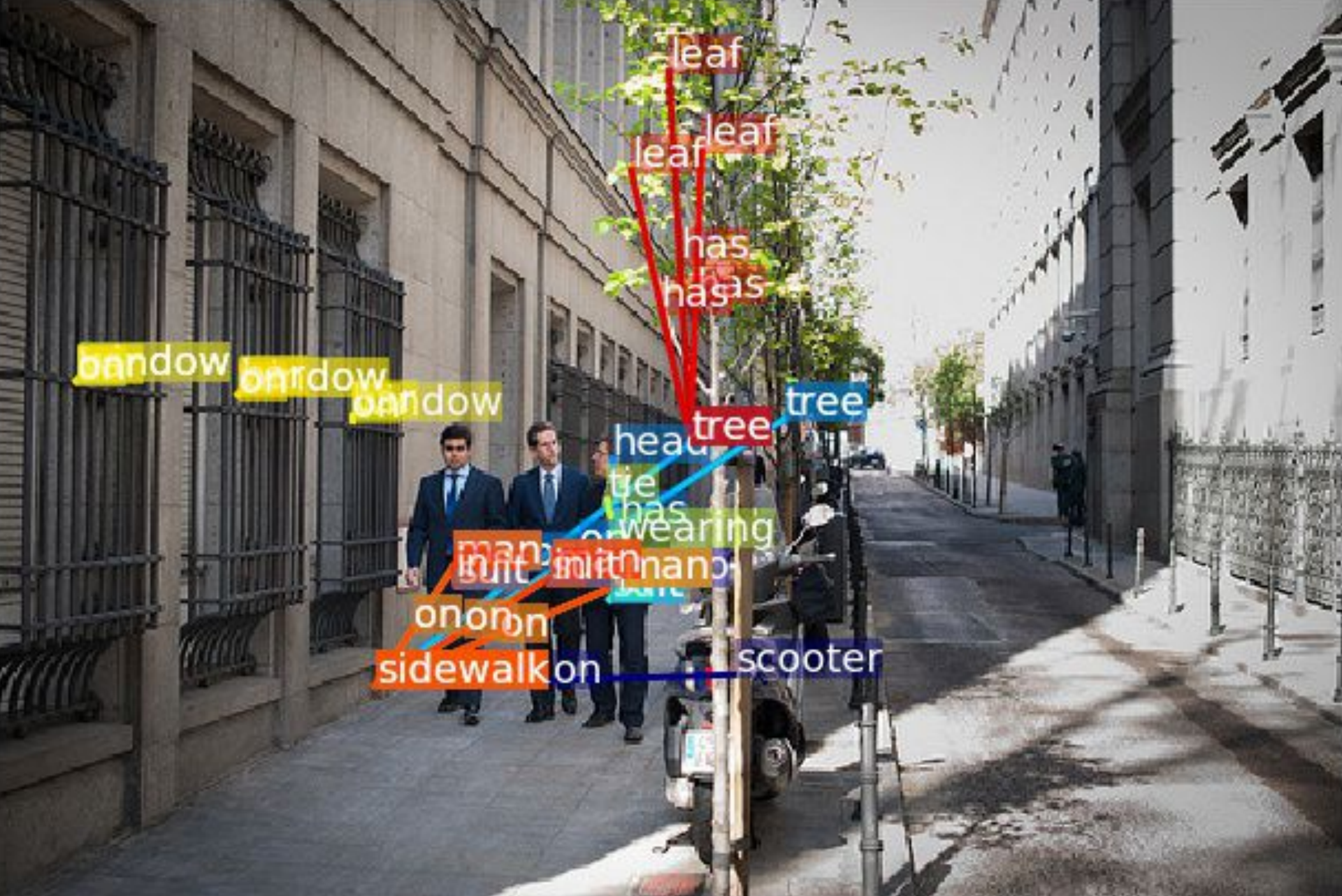}
    \end{subfigure}
  \end{subfigure}
  \begin{subfigure}{0.342\linewidth}
    \centering
    \begin{subfigure}{\linewidth}
      \centering
      \includegraphics[width=\textwidth]{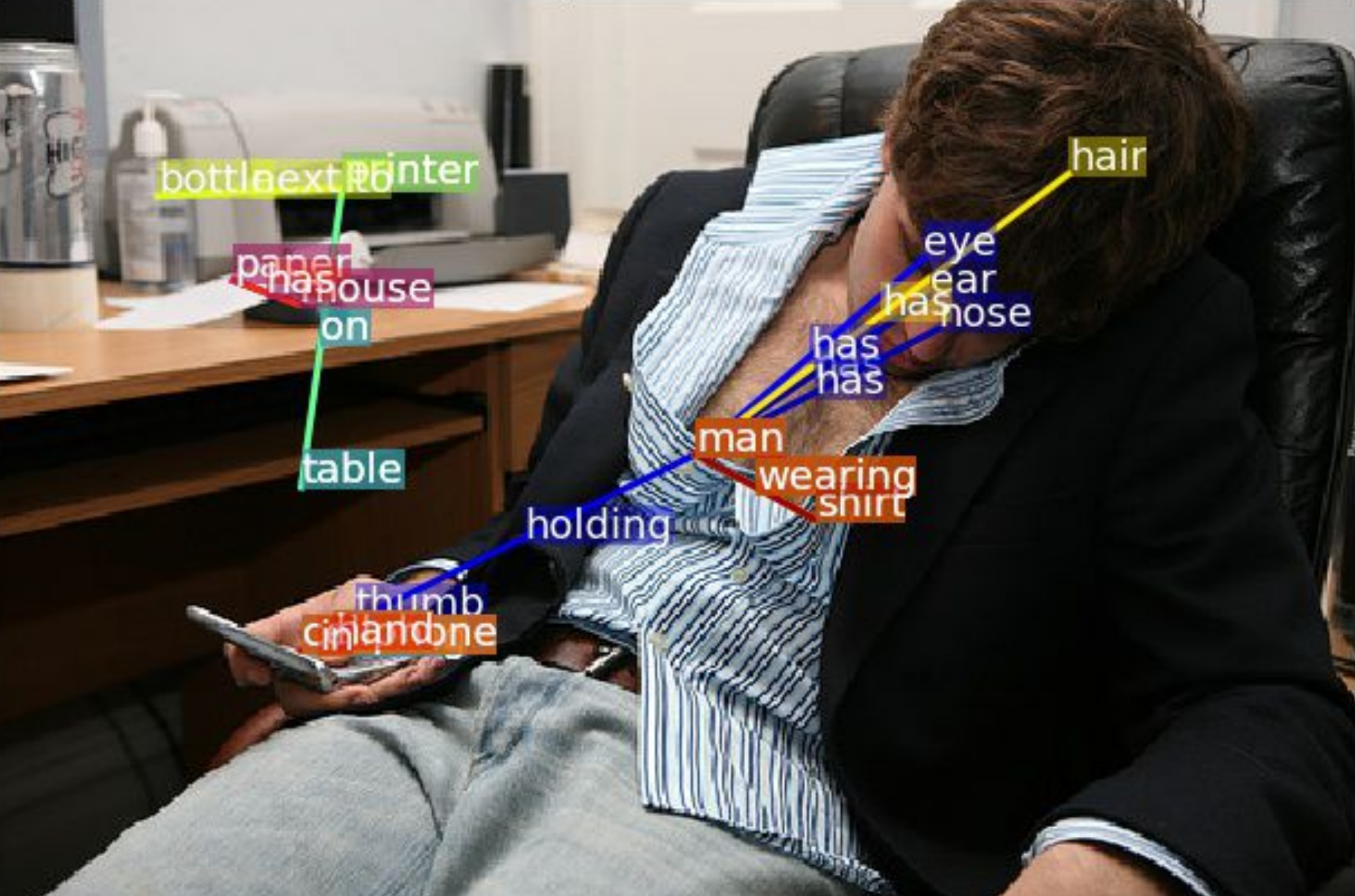}
    \end{subfigure}
  \end{subfigure}
  
  \centering
  \begin{subfigure}{0.338\linewidth}
    \centering
    \begin{subfigure}{\linewidth}
      \centering
	  \includegraphics[width=\textwidth]{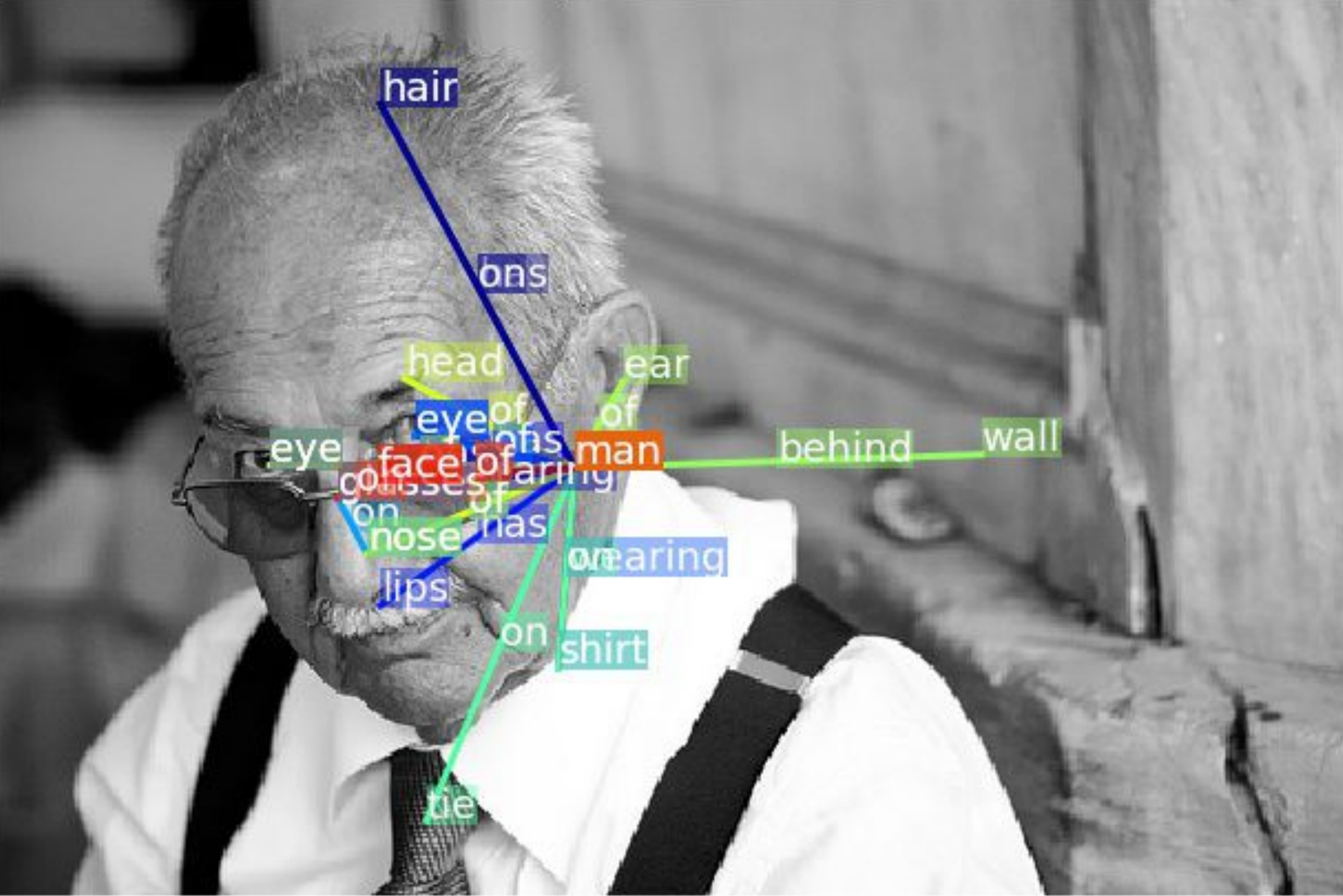}
    \end{subfigure}
  \end{subfigure}
  \centering
  \begin{subfigure}{0.328\linewidth}
    \centering
    \begin{subfigure}{\linewidth}
      \centering
      \includegraphics[width=\textwidth]{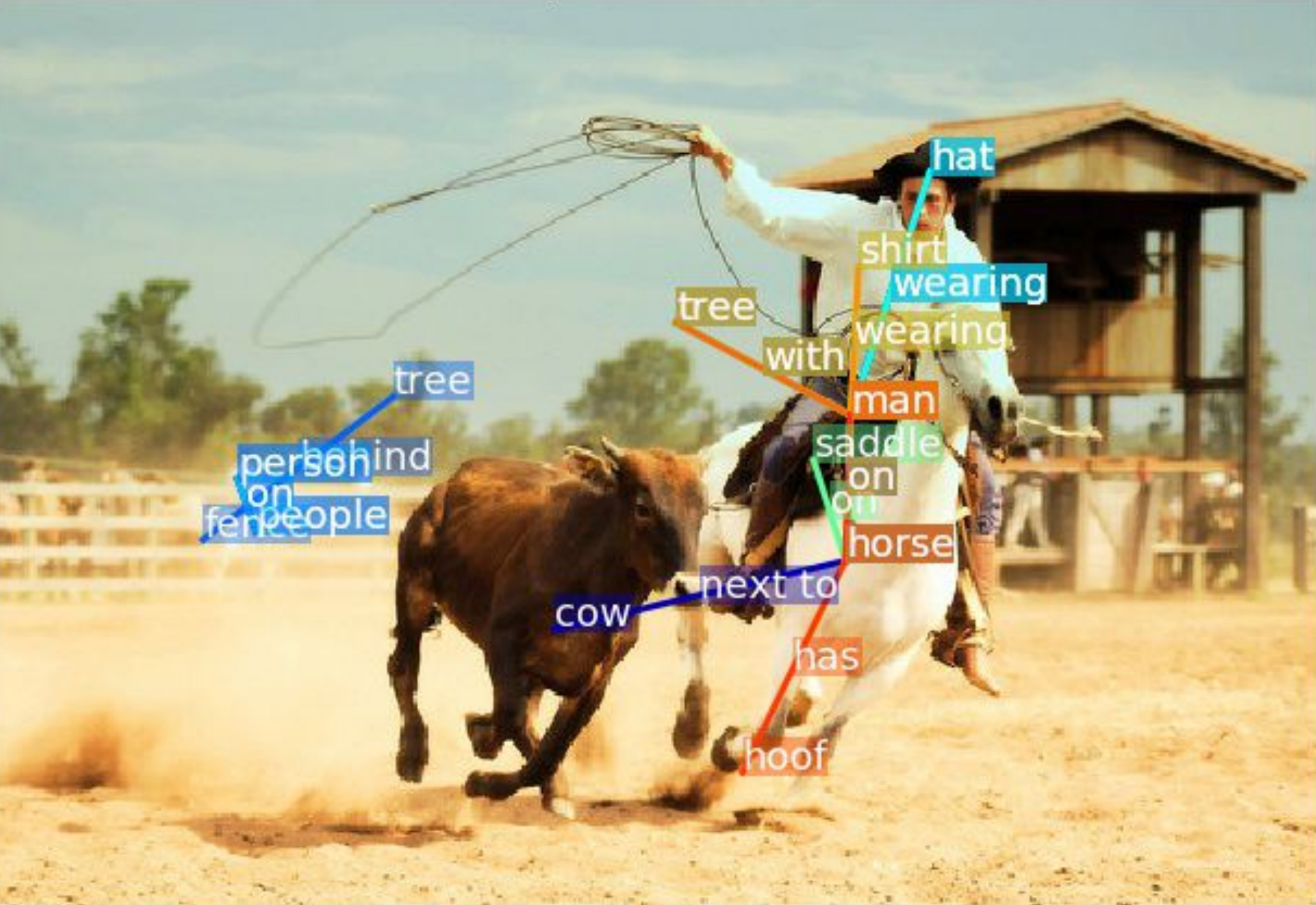}
    \end{subfigure}
  \end{subfigure}
  \begin{subfigure}{0.305\linewidth}
    \centering
    \begin{subfigure}{\linewidth}
      \centering
      \includegraphics[width=\textwidth]{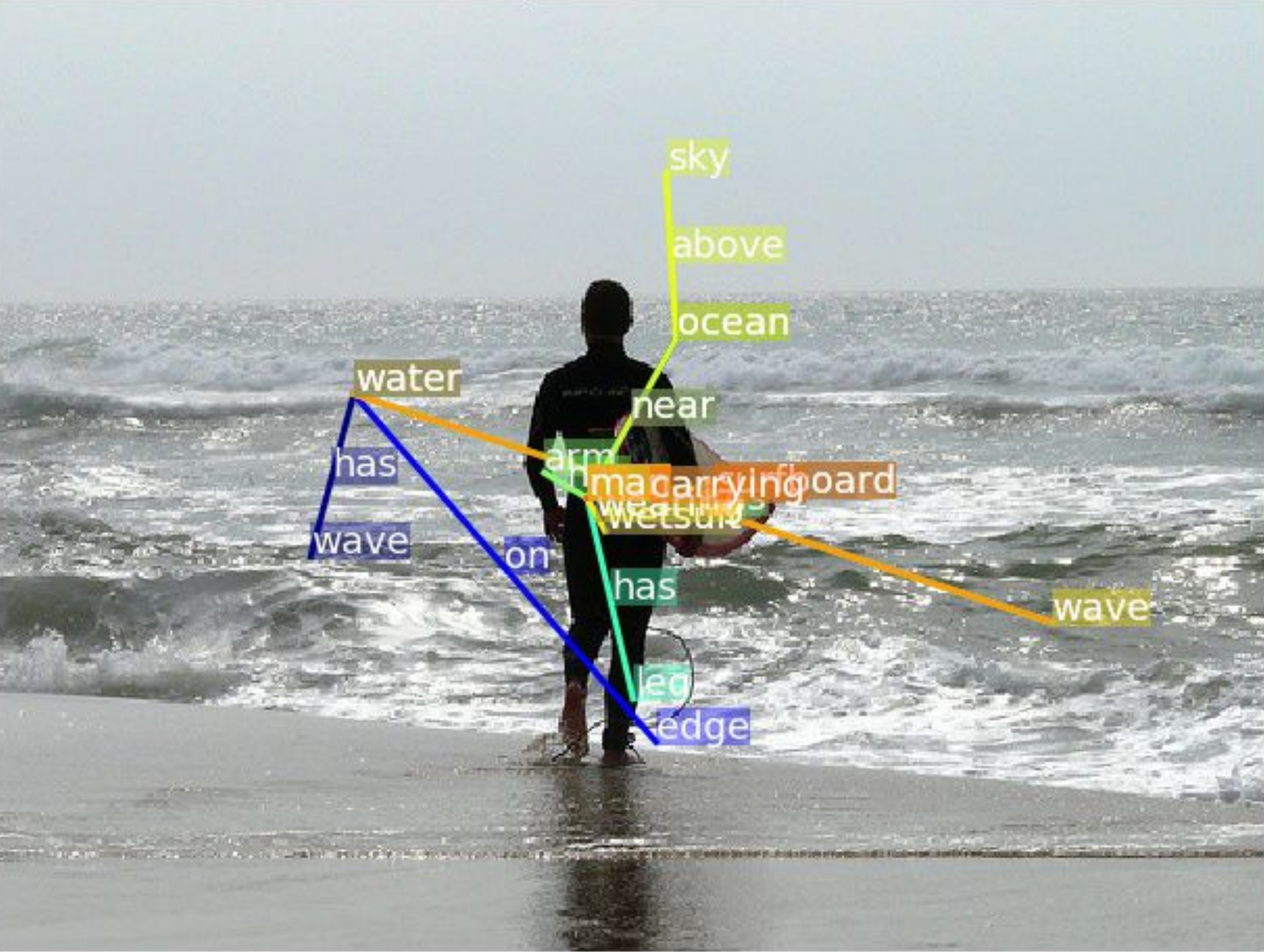}
    \end{subfigure}
  \end{subfigure}
\setlength\belowcaptionskip{-2ex}
\captionsetup{font=small}
\caption{Qualitative results. Our model recognizes a wide range of relation ship triples. Even if they are not always matching the ground truth they are frequently correct or at least reasonable as the ground truth is not complete.}
\label{fig:qualitative}
\end{figure*}

\head{The margin \textbf{m} in triplet loss}
We show results of triplet loss with various values for the margin $m$ in Table \ref{tab:triplet_m}. As described earlier, this value allows slackness in pushing negative pairs away from positive ones. We observe similar results with previous works \citep{kiros2014unifying,faghri2017vse++} that it is the best to set $m=0.1$  or $m=0.2$ in order to achieve optimal performance. It is clear that triplet loss is not able to learn discriminative embeddings that are suitable for classification tasks, even with larger $m$ that can theoretically enforce more contrast against negative labels. We believe that the main reason is that in a hinge loss form, triplet loss treats all negative pairs equally ``hard'' as long as they are within the margin $m$. However, as shown by the successful softmax models, ``easy'' negatives (\eg, those that are close to positives) should be penalized less than those ``hard'' ones, which is a property our model has since we utilize softmax for contrastive training.

\subsubsection{Qualitative results}
\label{subsubsec:qualitative}

The VG80k has densely annotated relationships for most images with a wide range of types. In Figure \ref{fig:qualitative} there are interactive relationships such as ``boy flying kite'', ``batter holding bat'', positional relationships such as ``glass on table'', ``man next to man'', attributive relationships such as ``man in suit'' and ``boy has face''. Our model is able to cover all these kinds, no matter frequent or infrequent, and even for those incorrect predictions, our answers are still semantic meaningful and similar to the ground-truth, e.g., the ground-truth ``lamp on pole'' v.s. the predicted ``light on pole'', and the ground-truth ``motorcycle on sidewalk'' v.s. the predicted ``scooter on sidewalk''.

%% file: conclusions.tex
\section{Conclusions}

In this work we study visual relationship detection at an unprecedented scale and propose a novel model that can generalize better on long tail class distributions. We find it is crucial to integrate subject and object features at multiple levels for good relation embeddings and further design a loss that learns to embed visual and semantic features into a shared space, where semantic correlations between categories are kept without hurting discriminative ability. We validate the effectiveness of our model on multiple datasets, both on the classification and detection task, and demonstrate the superiority of our approach over strong baselines and the state-of-the-art. 
Future work includes integrating a relationship proposal into our model that would enable end-to-end training.





